\documentclass[acmtog,authorversion]{acmartARXIV}

\usepackage{booktabs} 
\setcitestyle{square}

\usepackage[ruled]{algorithm2e} 

\SetAlFnt{\small}
\SetAlCapFnt{\small}
\SetAlCapNameFnt{\small}
\SetAlCapHSkip{0pt}
\IncMargin{-\parindent}

\acmJournal{TOG}

\setcopyright{rightsretained}

\acmDOI{}

\begin{document}
\title{FaceShop: Deep Sketch-based Face Image Editing}

\author{Tiziano Portenier}
\affiliation{%
  \institution{University of Bern}
  \streetaddress{Neubr\"uckstrasse 10}
  \city{Bern}
  \postcode{3012}
  \country{Switzerland}}
\email{portenier@inf.unibe.ch}

\author{Qiyang Hu}
\affiliation{%
  \institution{University of Bern}
  \streetaddress{Neubr\"uckstrasse 10}
  \city{Bern}
  \postcode{3012}
  \country{Switzerland}}
\email{hu@inf.unibe.ch}

\author{Attila Szab\'{o}}
\affiliation{%
  \institution{University of Bern}
  \streetaddress{Neubr\"uckstrasse 10}
  \city{Bern}
  \postcode{3012}
  \country{Switzerland}}
\email{szabo@inf.unibe.ch}

\author{Siavash Arjomand Bigdeli}
\affiliation{%
  \institution{University of Bern}
  \streetaddress{Neubr\"uckstrasse 10}
  \city{Bern}
  \postcode{3012}
  \country{Switzerland}}
\email{bigdeli@inf.unibe.ch}

\author{Paolo Favaro}
\affiliation{%
  \institution{University of Bern}
  \streetaddress{Neubr\"uckstrasse 10}
  \city{Bern}
  \postcode{3012}
  \country{Switzerland}}
\email{favaro@inf.unibe.ch}

\author{Matthias Zwicker}
\affiliation{%
  \institution{University of Maryland, College Park}
  \streetaddress{8223 Paint Branch Drive}
  \city{College Park}
  \state{MD}
  \postcode{20742}
  \country{USA}}
\email{zwicker@cs.umd.edu}

\renewcommand\shortauthors{Portenier, T. et al.}

\begin{abstract}
We present a novel system for sketch-based face image editing, enabling users to edit images intuitively by sketching a few strokes on a region of interest. Our interface features tools to express a desired image manipulation by providing both geometry and color constraints as user-drawn strokes. As an alternative to the direct user input, our proposed system naturally supports a copy-paste mode, which allows users to edit a given image region by using parts of another exemplar image without the need of hand-drawn sketching at all. 
The proposed interface runs in real-time and facilitates an interactive and iterative workflow to quickly express the intended edits.
Our system is based on a novel sketch domain and a convolutional neural network trained end-to-end to automatically learn to render image regions corresponding to the input strokes.
To achieve high quality and semantically consistent results we train our neural network on two simultaneous tasks, namely image completion and image translation. 
To the best of our knowledge, we are the first to combine these two tasks in a unified framework for interactive image editing. 
Our results show that the proposed sketch domain, network architecture, and training procedure generalize well to real user input and enable high quality synthesis results without additional post-processing.
\end{abstract}

%
%


%
%


\begin{teaserfigure}
  \includegraphics[width=\textwidth]{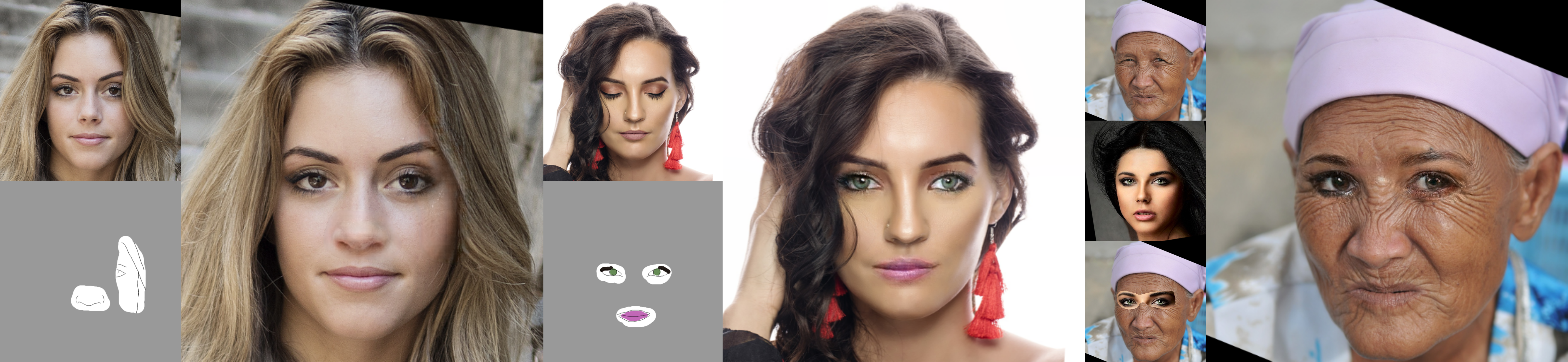}
  \caption{Results produced by our sketch-based face image editing system. For each example, we show the original image and the input to our system in the left column, and the final output in the right column. Left: changing nose and strand of hair using sketching. Middle: editing eyes, makeup, and mouth using sketching and coloring. Right: replacing eyes region using smart copy-paste. Photos from the public domain.}
  \label{fig:teaser}
\end{teaserfigure}

\maketitle

\section{Introduction}
Interactive image editing is an important field in both computer graphics and computer vision. Social media platforms register rapidly growing contributions of imagery content, which is creating an increasing demand for image editing applications that enable untrained users to enhance and manipulate photos. For example, Instagram reports more than 40~billion uploaded photos, and the number of photos is growing by 95~million per day~\cite{instastats}. However, there is a lack of tools that feature more complex editing operations for inexperienced users, such as changing the facial expression in an image. In this work, we propose a sketch-based editing framework that enables a user to edit a given image by sketching a few strokes on top of it. We show three examples produced by our system in Figure~\ref{fig:teaser} and more results in Section~\ref{sec:results}.

Image editing techniques can be divided into global and local editing operations. Recently, numerous deep learning techniques have been proposed for global editing applications such as general image enhancing~\cite{Yan2016,gharbi2017deep}, grayscale to color transformation~\cite{IizukaSIGGRAPH2016,zhang2017real}, and geometric, illumination, and style transformations~\cite{Selim2016,GatysEB15a,Kemelmacher2016,Liao2017}. Many of these global editing techniques build on image-to-image translation networks that learn to translate images from a source domain into a target domain~\cite{Isola_2017_CVPR,sangkloy2016scribbler,ZhuPIE17,Chen_2017_ICCV,wang2017high}. The synthesis quality of such networks has reached astonishing levels, but they lack generalizability to imperfect input and do not support local editing. While our work is technically related to these approaches, we focus more on local image editing operations and propose a source domain that generalizes well to hand-drawn user input.

Local image editing techniques manipulate only a limited spatial region of an input image, for example in the case of adding or removing objects in a scene, or changing their pose or shape. Techniques that are particularly successful in this direction solve an image completion problem to synthesize a missing region in an image, given the context as input. Recently, successful deep learning methods have been proposed to solve the image completion problem~\cite{pathakCVPR16context,Li_2017_CVPR,dolhansky2017eye,Yeh_2017_CVPR,IizukaSIGGRAPH2017}. A major issue with these approaches is that the synthesized contents are completely determined by the image context, and the user has no means to control the result. In this work, we also formulate local image editing operations as an image completion problem. However, we combine this approach with the aforementioned image translation approaches to enable fine-grained user control of the synthesis process.

An intuitive interface is crucial to enable users to express their intended edits efficiently, and interactivity is important to support an iterative workflow. Sketching has proven to be attractive to quickly produce a visual representation of a desired geometric concept, for example in image retrieval. Based on this observation, we propose a sketch-based interface for local image editing that allows users to constrain the shape and color of the synthesized result.

At the core of our framework is a novel formulation of conditional image completion. Given a spatial image region to be edited, the task is to replace this region by using the sketch-based user input and the spatial context surrounding the region. To solve this task we propose a novel sketch domain and training procedure for convolutional neural networks~(CNNs). More precisely, we train CNNs on two task jointly: image completion and image translation. In addition, our sketch domain naturally supports a smart copy-paste mode that allows a user to copy content from a source image and to blend it onto a target image. We show that this is very robust to illumination, texture, and even pose inconsistencies. Our proposed networks can be trained end-to-end on arbitrary datasets without labels or other prior knowledge about the data distribution.

We evaluate our approach on face image editing and compare our system to existing image completion and translation techniques. We show that our unified approach is beneficial because it enables fine-grained user control and is robust to illumination, texture, and pose inconsistencies. In summary, we make the following contributions:
\begin{itemize}
 \item the first end-to-end trained system that enables contextually coherent, high-quality, and high-resolution local image editing by combining image completion and image translation,
 \item a very stable network architecture and training procedure,
 \item a technique to generate 
training sketches from images that enjoys a high generalizability to real user input and enables sketch-based editing with color constraints and smart copy-paste in a unified framework,
 \item globally consistent, seamless local edits without any further post-processing, and
 \item an intuitive sketch-based, interactive interface that naturally enables an efficient and iterative workflow.
\end{itemize}

\section{Related Work}
Interactive image editing approaches have a long history in computer graphics,
and providing a comprehensive survey exceeds the scope of this work. In this section, we discuss existing techniques that are highly related to our work. We group related works in two main categories: image completion and image translation. Image completion (also known as image inpainting) is the task of completing missing regions in a target image given the context. We tackle the problem of sketch-based local image editing by solving a conditional variant of the image completion problem. On the other hand, image translation considers the problem of transforming entire images from a source to a target domain, such as transforming a photo into an artistic painting, or transforming a semantic label map into a photo-realistic image. In our work, we consider transforming a sketch-based user input into a photographic image patch that replaces the region to be edited in the target image.

\subsection{Image Completion}
In their groundbreaking work, Bertalmio et al.~\shortcite{Bertalmio:2000:II} were inspired by techniques used by professional image restorators, and their procedure attempts to inpaint missing regions by continuing isophote lines. Shen and Chan~\shortcite{Shen:2001:MML} built on this work and connected it to variational models. Patch-based procedures are a popular alternative: given a target image with a missing region to be filled-in, the idea is to use image patches occurring in the context surrounding the missing region, or in a dedicated source image, to replace the missing region. Pioneering work by Efros and Leung~\shortcite{Efros2001} leverages this strategy for texture synthesis, and various researchers later extended it for image completion~\cite{drori2003fragment,bertalmio2003simultaneous,criminisi2004region}. PatchMatch~\cite{Barnes2009} accelerates the patch search algorithm to enable image completion at interactive frame rates, and Image Melding~\cite{Darabi2012} further improves the quality of the patch search algorithm. A major issue with patch-based approaches is that they work by copying and pasting pixel values from existing examples. Therefore, they rely on suitable example patches in the context and they cannot invent new objects or textures.

\begin{figure*}[t]
\centering
\includegraphics[width=1\textwidth]{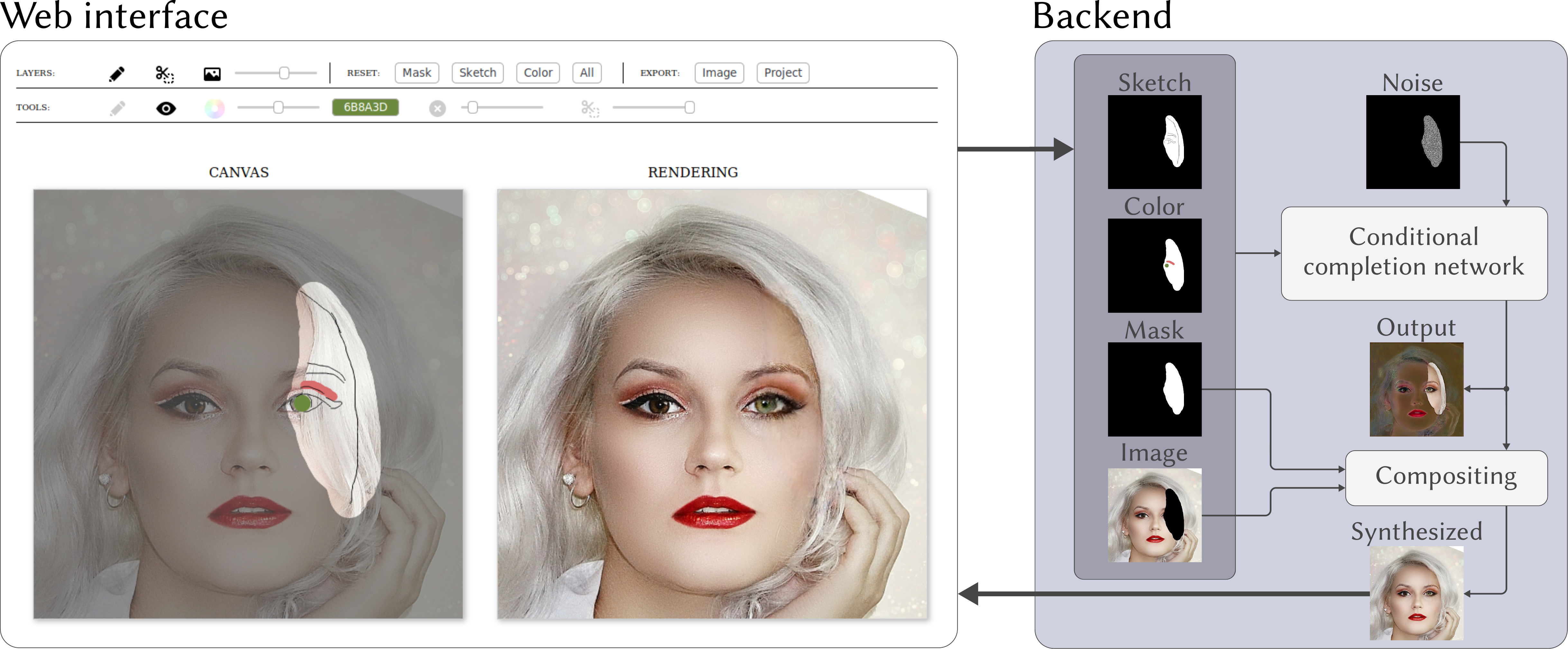}
\caption{System overview. The web-based interface allows users to specify masking regions, an edge sketch, and color strokes. The core of the backend is a conditional image completion network that takes the user input, the original image, and a noise layer, and reconstructs a full image. The reconstructed image is composited with the original by filling the masked region in the original with the reconstructed output, and returned. Photo from the public domain.}
\label{fig:system_overview}
\end{figure*}

Recently, methods based on CNNs have been proposed to overcome this issue. Training neural networks for image completion enables systems that can generate content for missing regions, based on the data that they have seen during training. Using a pixel-wise reconstruction loss for training such networks leads to contextually consistent, but blurry results, because the problem is highly ambiguous. Context Encoders~\cite{pathakCVPR16context} mitigate this problem by training Generative Adversarial Networks~(GANs)~\cite{goodfellow2014}. GANs consist of an auxiliary discriminator network, which is trained to distinguish synthesized from real data, and a synthesis network, which is trained to fool the discriminator by generating high-quality synthetic data. This approach proved to be successful for image completion tasks, and several systems have been proposed along this line. In particular, Dolhansky et al.~\shortcite{dolhansky2017eye} apply this concept on the task of eye inpainting to replace closed eyes in face images, given an exemplar image of the same identity. Yeh et al.~\shortcite{Yeh_2017_CVPR} proposed an iterative optimization technique to complete images using adversarial training. Further improvements of synthesis quality have been achieved by using multiple discriminator networks for different scales~\cite{IizukaSIGGRAPH2017,Li_2017_CVPR}.

Most image completion approaches discussed above have in common that they condition the synthesis process only on the context of the image to be completed. The user has no further control on the contents that are synthesized and the completion result is largely determined by the training data. The only exception is the eye inpainting technique by Dolhansky et al.~\shortcite{dolhansky2017eye}. However, their system can only complete missing eye regions. In our sketch-based image editing system we also leverage a GAN loss to train CNNs for image completion. In contrast to the techniques discussed above, however, we formulate a conditional image completion problem and the user can guide the geometry and color of the synthesized result using sketching. Finally, it must be mentioned that Poisson image editing~\cite{Perez2003} can also be considered as a technique for conditioned image completion. Thus, we compare it to our sketch-domain approach and show that our approach is more robust to illumination, texture, or shape inconsistencies.

\subsection{Image-to-Image Translation}
Several techniques have been proposed in the past to translate the semantic and geometric content of a given image into a target domain with different style~\cite{GatysEB15a,Selim2016,Kemelmacher2016}. For example, Gatys et al.~\shortcite{GatysEB15a} trained a network that turns a source photo into paintings of famous artists, given an exemplar painting. The synthesis is an optimization procedure constrained to match the style of a given exemplar painting by matching deep feature correlations of the exemplar, and content is preserved by matching deep features of the source image. Recently, image-to-image translation networks have been proposed that translate an image from a source domain into a target domain using a single feedforward pass~\cite{Isola_2017_CVPR,sangkloy2016scribbler,ZhuPIE17}. For example, the network by Isola et al. translates images from a source domain, for example edge maps or semantic label maps, into photos. The idea is to train a conditional variant of GANs~\cite{MirzaO14} that encourages the synthesized result to correspond to auxiliary input information. Chen and Koltun~\shortcite{Chen_2017_ICCV} propose a cascaded multi-scale network architecture to improve the visual quality and increase spatial resolution of the translated images. In our work, we also train conditional GANs in an image translation fashion to achieve sketch-based image editing, but we formulate it as a completion problem.

\section{Deep Sketch-based Face Image Editing}
In this section we introduce our sketch-based image editing system, see Figure~\ref{fig:system_overview} for an overview. The system features an intuitive and interactive user interface to provide sketch and color information for image editing. The user input is fed to a deep CNN that is trained on conditional image completion using a GAN loss. Image completion proceeds at interactive rates, which allows a user to change and adapt the input in an efficient and iterative manner. The core component of our system is a CNN that incorporates both image context and sketch-based user input to synthesize high-quality images, unifying the concept of image completion and image translation. 

We discuss our training data and propose a suitable sketch domain in Section~\ref{subsec:data}. Using appropriate training data is crucial to produce convincing results with real user input. In Section~\ref{subsec:net_arch} we describe a network architecture that enables the synthesis of high-quality and high-resolution images. Then, in Section~\ref{subsec:training} we discuss a training procedure that allows us to train our model end-to-end with fixed hyper-parameters in a very stable manner. Finally, in Section~\ref{subsec:ui} we discuss in detail our sketch-based user interface.

\subsection{Training Data}
\label{subsec:data}
In order to train our model on conditional image completion, we generate training data by removing rectangular masks of random position and size from the training images (Section~\ref{sec:masking}). To obtain a system that is sensitive to sketch-based user input, we provide additional information to the network for the missing region. The design of an appropriate sketch domain (Section~\ref{sec:sketch-domain}) is crucial to achieve convincing results with hand-drawn user input. In addition, our system also includes color data (Section~\ref{sec:color-domain}).

\subsubsection{Rectangular Masking}
\label{sec:masking}
We first experimented with axis-aligned rectangular editing regions during training, similar to previous image completion methods~\cite{pathakCVPR16context,IizukaSIGGRAPH2017}. However, we observe that the model overfits 
to axis-aligned masks and the editing results exhibit distracting visible seams if the region to be edited is not axis-aligned. Hence we rotate rectangular editing regions during training by a random angle $\alpha \in [0, 45]$ to teach the network to handle all possible slopes. In Section~\ref{subsec:ablation} we show that this extension effectively causes the model to produce seamless editing results for arbitrarily shaped editing regions.

\subsubsection{Sketch Domain}
\label{sec:sketch-domain}

On the one hand, it is beneficial to transform training imagery automatically into the sketch domain to quickly produce a large amount of training data. On the other hand, the trained system should generalize to real user input that may deviate significantly from automatically generated data. Recent work in image translation has shown that automatically generated edge maps produce high-quality results when translating from edges to photos~\cite{Isola_2017_CVPR,sangkloy2016scribbler}. However, these models tend to overfit to the edge map domain seen during training, and it has been shown that the output quality of these models decreases when using hand-drawn sketches instead of edge maps~\cite{Isola_2017_CVPR}. To mitigate this problem we propose an automatic edge map processing procedure that introduces additional ambiguity between the input sketch domain and the ground truth translated result, and we show in Section~\ref{subsec:ablation} that this approach increases the network's ability to generalize to hand-drawn input after training. 

\begin{figure}[t]
\centering
\includegraphics[width=1\linewidth]{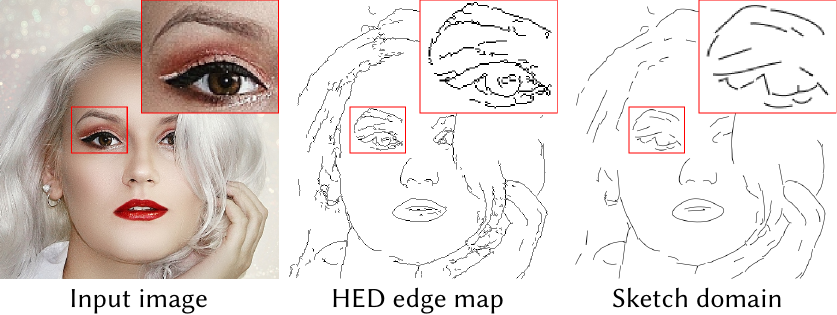}
\caption{Sketch domain. We extract edge maps using HED~\cite{Xie_2015_ICCV}, and fit splines using AutoTrace~\cite{autotrace}. We then remove small edges, smooth the control points, and rasterize the curves. Photo from the public domain.}
\label{fig:sketch_prep}
\end{figure}
Figure~\ref{fig:sketch_prep} shows an example of our proposed sketch domain transform. We first extract edge maps using the HED edge detector~\cite{Xie_2015_ICCV}, followed by splines fitting using AutoTrace~\cite{autotrace}. After removing small edges with a bounding box area below a certain threshold, we smooth the control points of the splines. This step is crucial, since the smoothed curves may now deviate from the actual image edges. This introduced ambiguity encourages the model not to translate input strokes directly to image edges, but to interpret them as rough guides for actual image edges while staying on the manifold of realistic images. After these preprocessing steps we rasterize the curves. This final step completes the transformation into the proposed sketch domain. 

\subsubsection{Color Domain}
\label{sec:color-domain}
\begin{figure}[t]
\centering
\includegraphics[width=1\linewidth]{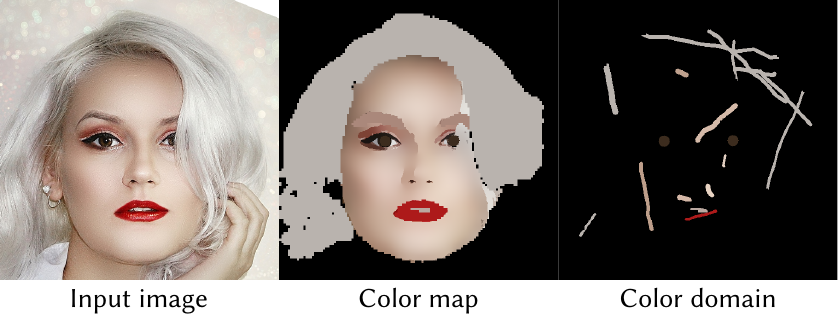}
\caption{Color domain. We pre-process the input image using dowsampling and filtering, and generate random color strokes to resemble user input. Photo from the public domain.}
\label{fig:color_domain}
\end{figure}

In order to enable a user to constrain the color of the edited result, we provide additional color information for the missing region to the networks during training. An intuitive manner for a user to provide color constraints is to draw a few strokes of constant color and arbitrary thickness on top of the image. For this purpose we propose a technique to automatically transform a training image into a color map that is suitable to generate a random color strokes representation for training, see Figure~\ref{fig:color_domain} for an example. Our approach to produce random color strokes has some similarities to the method proposed by Sangkloy et al.~\shortcite{sangkloy2016scribbler}. However, our novel color map transformation introduces further ambiguities to make the system more robust to real user input. Moreover, in the case of face images we also leverage semantic information to further improve the result.

We first downsample the input image to $128 \times 128$ and apply a median filter with kernel size 3. Next, we apply bilateral filtering with $\sigma_r = 25$ and $\sigma_d = 7$ 40 times repeatedly. This results in a color map with largely constant colors in low-frequency regions, which mimes the intended user input. In the case of face images we increase the ambiguity in the color map using semantic information. We first compute semantic labels using the face parsing network~\cite{Li_2017_CVPR}. Next, we compute median colors for hair, lips, teeth, and eyebrows regions and replace the corresponding regions in the color map with these median colors. This allows a user later to quickly change e.g. lips color with a single, constantly colored stroke.

During training, we generate a random number of color strokes by randomly sampling thickness and start and end points of the strokes and interpolate between these points with additional random jittering orthogonal to the interpolation line. We color the random strokes using the color map value of the start position. If the color map value at the current stroke point position deviates more than a certain threshold from the initial value, we terminate the current stroke early. This technique results in random color strokes of constant color that may deviate from but correlate with the actual color in the input image.

In our experiments we observed that the model sometimes produces inconsistent iris colors. For example, the network sometimes renders two differently colored eyes. This observation is consistent with the findings by Dolhansky and Ferrer~\shortcite{dolhansky2017eye} and we propose to mitigate this issue using iris detection. We first detect pupils using the work by Timm and Barth~\shortcite{timm2011accurate}. After normalizing the image size given the eye bounding box, we compute median iris colors in a fixed-size circle centered at the eye. This approach yields accurate iris colors in most cases and we replace the color map values at the iris positions using this color. During training, we draw a fixed-size circle of 10 pixels radius at the pupil positions, as opposed to the stroke-based approach used for all other color constraints. Our system should produce meaningful results even without additional color constraints, and only incorporate color information to guide the editing if available. Therefore, we provide color information during training only with probability $0.5$ for each image and the model has to decide on an appropriate color for the rest of the samples.

\begin{figure}[t]
\centering
\includegraphics[width=1\linewidth]{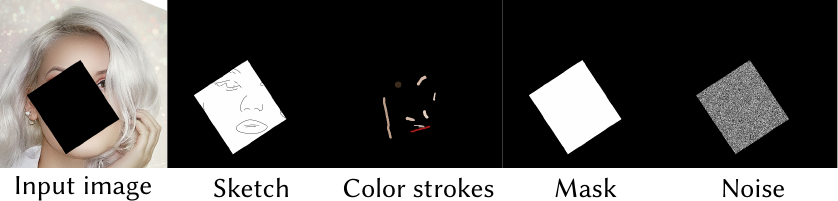}
\caption{The input to the conditional completion network includes the original image (with the masked region removed), edge sketch, color strokes, mask, and a noise layer. Photo from the public domain.}
\label{fig:generator_input}
\end{figure}

In addition to the sketch and color constraints we also feed per-pixel noise to the network, which enables the model to produce highly detailed textures by resolving further ambiguities, given an appropriate network architecture. Figure~\ref{fig:generator_input} summarizes the training data that we use as input to the conditional completion network.

\subsubsection{High-resolution Dataset}
To create our high-resolution face image dataset, we start with the in-the-wild images from the celebA dataset~\cite{liu2015facea}. We first remove all images with an annotated face bounding box smaller than $300 \times 300$ pixels. We roughly align the remaining 21k images by rotating, scaling, and translating them based on the eye positions. Images that are smaller than $512 \times 512$ pixels are padded with the image average color. Finally, we center crop all images to $512 \times 512$ pixels. We use 20k images for training and the remaining 1k images for testing.

\subsection{Network Architecture}
\label{subsec:net_arch}
Inspired by the recent work on deep image completion~\cite{IizukaSIGGRAPH2017}, our conditional completion network relies on an encoder-decoder architecture and two auxiliary discrimination networks. We tried out several architectural choices, using both a smaller cropped celebA dataset and our high resolution dataset. Our final architecture described in detail below, is very stable for training and produces high-quality synthesis results of up to $512 \times 512$ pixels. 

Training a neural network with a GAN loss involves the simultaneous training of two networks: the conditional image completion network, also called the \textit{generator} in GAN terminology, and the auxiliary \textit{discriminator}. 
The discriminator takes as input either real data or fake data produced by the generator network. It is trained to distinguish real from fake data, while the generator is trained to produce data that fools the discriminator. After the training procedure, the discriminator network is not used anymore, it only serves as a loss function for the generator network. 

In our image editing context the discriminator tries to distinguish edited photos from genuine, unmodified images. To encourage the conditional completion network not to ignore the additional sketch and color constraints, we provide this information also to the discriminator network as additional source to distinguish real from fake data. Because of the construction of the training data, real data always correlates with the sketch and color constraints. Therefore, training with this conditional GAN loss~\cite{MirzaO14} forces the conditional completion network to output data that correlates with the input. 

\subsubsection{Conditional Completion Network}
\begin{figure}[t]
\centering
\includegraphics[width=1\linewidth]{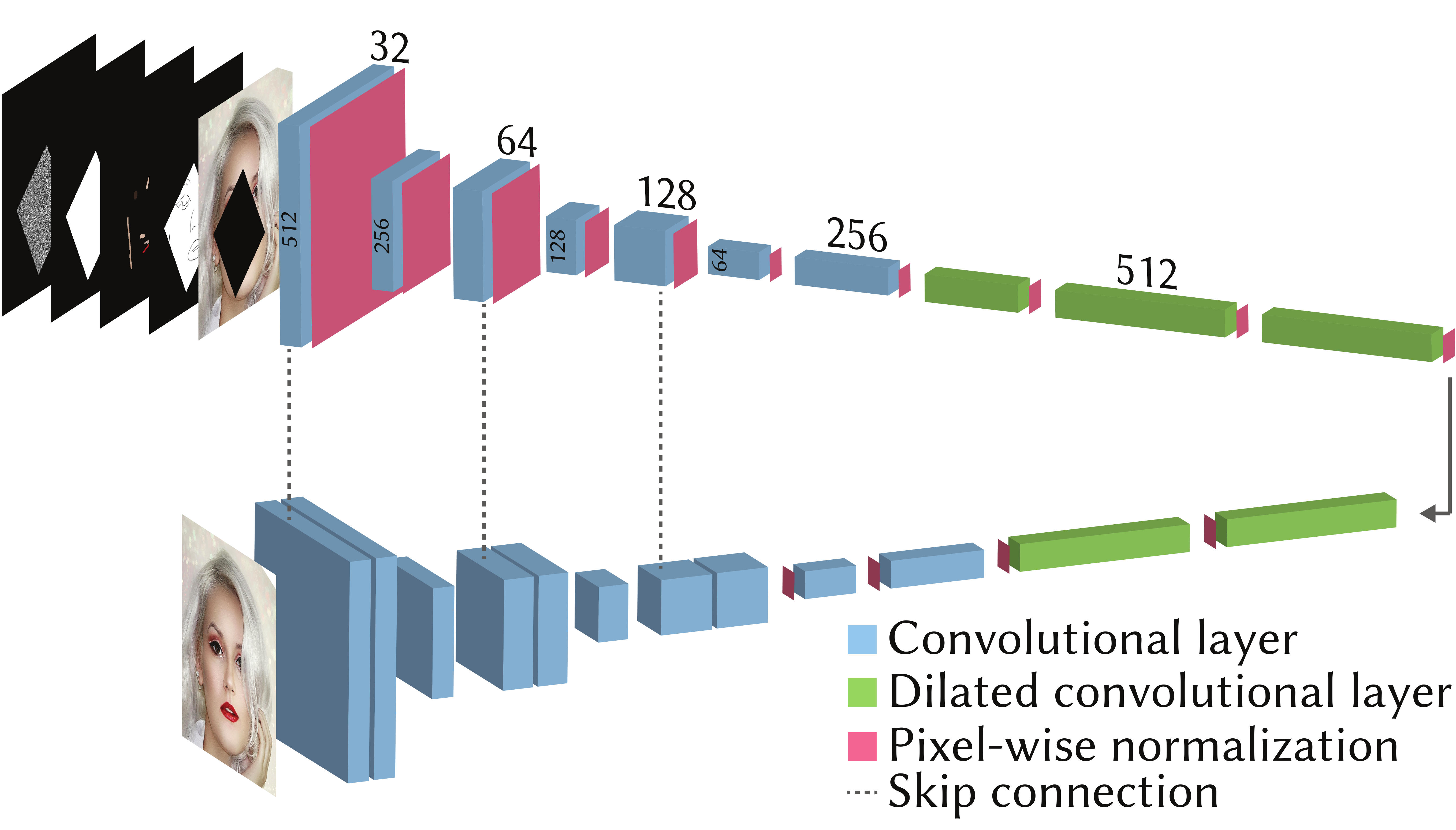}
\caption{Conditional completion network. Photo from the public domain.}
\label{fig:gen_arch}
\end{figure}
Figure~\ref{fig:gen_arch} shows a visualization of our proposed architecture for the conditional completion network. The input to our network is a tensor of size $512 \times 512 \times 9$: an input RGB image with a region to be edited removed, a binary sketch image that describes the shape constraints of the intended edit, a potentially empty RGB image that describes the color constraints, a binary mask that indicates the region to be edited, and a one-dimensional per-pixel noise channel. Figure~\ref{fig:generator_input} shows an example training input $I_g$ to the conditional completion network. The output of the network is an RGB image of size $512 \times 512$, and we replace the image context outside the mask with the input image before feeding it to the loss function (composition step in Figure~\ref{fig:system_overview}). This guarantees that the editing network is not constrained on image regions that are outside the editing region and the system enables local edits without changing other parts of the image.

\begin{figure*}[t]
\centering
\includegraphics[width=1\textwidth]{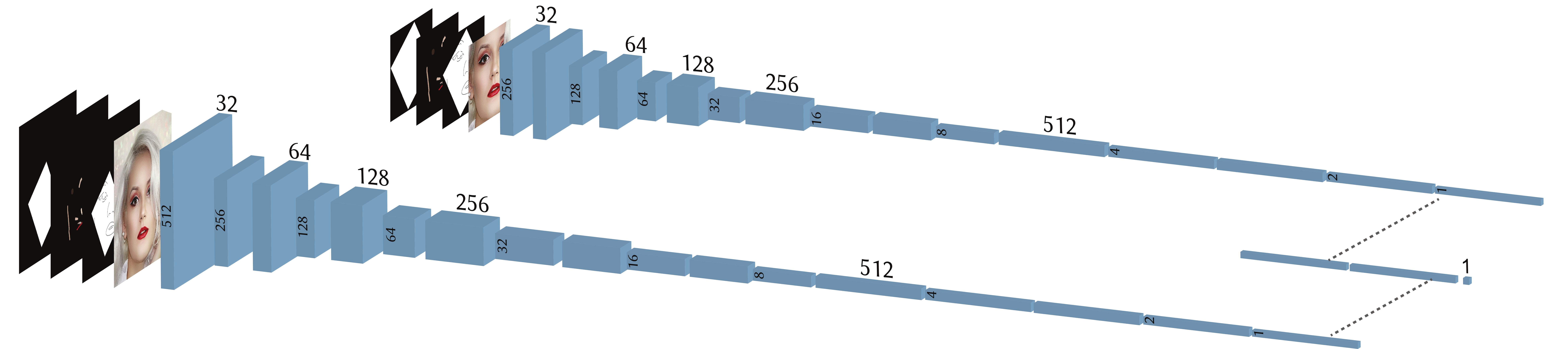}
\caption{Discriminator architecture consisting of a local (top) and global (bottom) network. Photo from the public domain.}
\label{fig:dis_arch}
\end{figure*}

We build on the encoder-decoder architecture proposed by Iizuka et al.~\shortcite{IizukaSIGGRAPH2017}, and add more layers and channels to reach the target image size of $512 \times 512$ pixels. Our final fully-convolutional high-resolution model downsamples the input image three times using strided convolutions, followed by intermediate layers of dilated convolutions~\cite{YuKoltun2016} before the activations are upsampled to $512 \times 512$ pixels using transposed convolutions. After each upsampling layer we add skip connections to the last previous layer with the same spatial resolution by concatenating the feature channels. These skip connections help to stabilize training on the one hand, and they allow the network to synthesize more realistic textures on the other hand. We observe that without these skip connections the network ignores the additional noise channel input and produces lower quality textures, especially for hair textures. 

Similar to~\cite{karras2017}, we implement a variant of local response normalization (LRN) after convolutional layers, defined as
\begin{equation}
\label{eq:lrn}
LRN(a_{x,y}) = \frac{a_{x,y}}{\sqrt{\frac{1}{N}\sum_{i=0}^{N-1}(a_{x,y}^i)^2 + \epsilon}},
\end{equation}
where $a_{x,y}^i$ is the activation in feature map $i$ at spatial position $(x,y)$, $N$ is the number of feature maps outputted by the convolutional layer, and $\epsilon = 10^{-8}$. We find this per-layer normalization in the editing network crucial for stable training, however, we also observe that this heavy constraint limits the capacity of the network dramatically and prevents the model from producing detailed textures. Our proposed solution is to apply LRN only after the first 14 layers before upsampling the data, which leads to both stable training and high-quality texture synthesis. We use the leaky ReLU activation function after each layer except for the output layer, which uses a linear activation function. In total, the proposed editing network consists of 23 convolutional layers with up to 512 feature channels.

\subsubsection{Discriminator Networks}

Figure~\ref{fig:dis_arch} shows a visualization of our proposed architecture for the discriminator network. Similar to Iizuka et al.~\shortcite{IizukaSIGGRAPH2017}, we use a global and a local discriminator. The input to the global network is a $512 \times 512 \times 8$ tensor: a fake sample $I_f$ consisting of the edited RGB image synthesized by the conditional completion network, sketch and color constraints, and the binary mask indicating the edited region. For real samples $I_r$ we use a genuine, unedited image and a random mask. The local discriminator uses the same input tensor but looks at a cropped region of size $256 \times 256$ centered around the region to be edited. The outputs of both discriminators are 512-dimensional feature vectors that are concatenated and fed into a fully-connected layer that outputs a single scalar value. This way the contribution of both discriminators is optimally weighted by learning the weights for the fully-connected layer and there is no need for a hyper-parameter.

Both discriminators are fully-convolutional with alternating convolution and strided convolution layers until the activations are downsampled to a single pixel with 512 feature channels. We use leaky ReLU activation functions everywhere in the discriminators except for the fully-connected output layer that uses a linear activation function. We do not apply any form of normalization in the discriminator networks. The global discriminator has 17 convolutional layers and the local discriminator consists of 16 layers.

\subsection{Training Procedure}
\label{subsec:training}

We experimented with several loss functions to achieve stable training behavior and high-quality synthesis results. Previous work has shown that a combination of pixel-wise reconstruction loss and GAN loss results in high-quality image synthesis in both image completion and image translation applications~\cite{pathakCVPR16context,IizukaSIGGRAPH2017,Isola_2017_CVPR,wang2017high}. The pixel-wise reconstruction loss stabilizes training and encourages the model to maintain global consistency on lower frequencies. We use $L_1$ reconstruction loss restricted to the area to be edited, that is
\begin{equation}
\label{eq:l1}
L_{rec}(x') = \frac{1}{N}\sum_{i=0}^{N-1}|x_i' - x_i|,
\end{equation}
where $x_i'$ is the output of the editing network at pixel $i$ after restoring the ground truth context, and $N$ is the number of pixels.

For the GAN loss we tried out three different approaches until we found a setting that works reasonably well for our system. We first experimented with the original GAN loss formulation by Goodfellow et al.~\shortcite{goodfellow2014}, that is
\begin{equation}
\label{eq:gan}
\min_{G} \max_{D} = \mathbb{E}[\log D(I_r)] + \mathbb{E}[log(1-D(I_f))],
\end{equation}
where $D$ is the discriminator and $G$ the generator network. However, we find that this GAN loss results in very unstable training and the networks often diverge, even for smaller networks on lower resolution data. Next we evaluated the BEGAN loss function~\cite{berthelotSM17} by replacing our discriminator networks with autoencoders. While this results in significantly more stable training behavior, the BEGAN setting converges very slowly and tends to produce more blurry synthesis results compared to the original GAN formulation. Moreover, the BEGAN discriminators occupy significantly more memory due to the autoencoder architecture and therefore limit the capacity of our model for high-resolution images due to memory limitations on current GPUs.

We achieve by far the best results on both high and low-resolution data using the WGAN-GP loss function~\cite{gulrajani2017} defined as
\begin{equation}
\label{eq:wgangp}
L_{WGAN-GP} = \mathbb{E}[D(I_f)] - \mathbb{E}[D(I_r)] + \lambda \mathbb{E} [(||\nabla_{I_u}D(I_u)||_2 - 1)^2],
\end{equation}
where $\lambda$ is a weighting factor for the gradient penalty and $I_u$ is a data point uniformly sampled along the straight line between $I_r$ and $I_f$. Similar to Karras et al.~\shortcite{karras2017} we add an additional term $\epsilon \mathbb{E}[D(I_r)^2]$, to keep the discriminator output close to zero. Our overall loss function therefore becomes
\begin{equation}
\label{eq:loss}
L = \alpha L_{rec} + L_{WGAN-GP} + \epsilon \mathbb{E}[D(I_r)^2],
\end{equation}
and we set $\alpha = \epsilon = 10^{-3}$ and $\lambda = 100$. We use a learning rate of $2 \times 10^{-4}$ without any decay schedule and train using the ADAM optimizer~\cite{kingmaB14} with standard parameters. With this hyper-parameter setting we are able to train our high-resolution model end-to-end without any parameter adjustments during training, unlike Iizuka et al.~\shortcite{IizukaSIGGRAPH2017}. We attribute this stability to the WGAN-GP loss and our network architecture, mainly the local response normalization and the skip connections in the conditional completion network. Training the full model takes two weeks on a Titan XP GPU with batch size of 4.

\subsection{User Interface}
\label{subsec:ui}
Our web-based user interface (Figure~\ref{fig:system_overview} left) features several tools to edit an image using sketching. The interface consists of two main canvases: the left canvas shows the original input image and a user can sketch on it to edit. The right canvas shows the edited result and is updated immediately after drawing each stroke. We provide a mask tool to draw an arbitrary shaped region to be edited. A user can indicate the shape of the edited contents by drawing and erasing strokes using the pen tool. A color brush tool allows to draw colored strokes of variable thickness to indicate color constraints. For eye colors the interface provides a dedicated iris color tool that allows to draw colored circles to indicate iris color and position. A forward pass through the conditional completion network takes 0.06 seconds on the GPU, which allows a real-time user experience.

\section{Results}
\label{sec:results}

We next present ablation studies to demonstrate the benefits of various components of our system (Section~\ref{subsec:ablation}), followed by image editing results (Sections~\ref{sec:sketch-based-face-image-editing}, \ref{sec:smart-copy-paste}) and comparisons to related techniques (Section~\ref{sec:comparisons}).

\subsection{Ablation Studies}
\label{subsec:ablation}

\begin{figure}[t]
\centering
\includegraphics[width=1\linewidth]{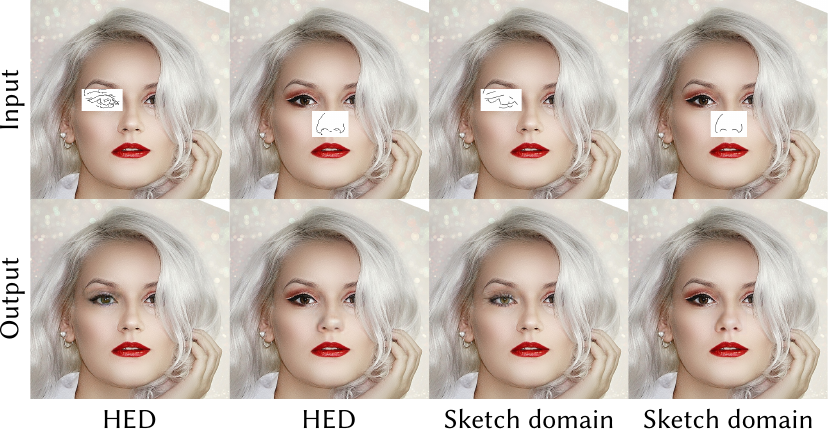}
\caption{Editing results on training-like data for HED edges and our proposed sketch domain. Both lead to high quality conditional image completion. Photo from the public domain.}
\label{fig:training_like_hed_vs_hedprep}
\end{figure}

\begin{figure}[t]
\centering
\includegraphics[width=1\linewidth]{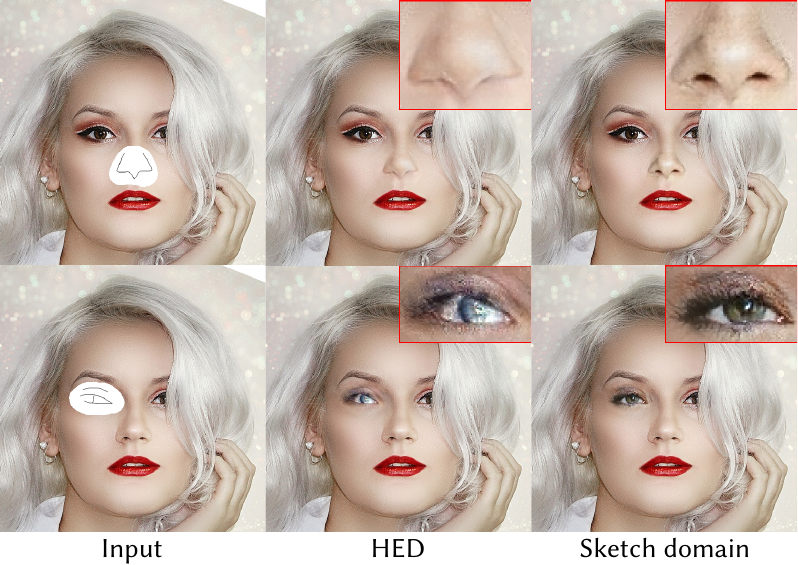}
\caption{Editing results on real user input for a system trained using HED edges (middle) and our proposed sketch domain (right). Using HED edges, artifacts in the user input translate into artifacts in the output. Our sketch domain leaves some ambiguity between image structures and sketches, allowing to suppress inaccuracies in the input. Photo from the public domain.}
\label{fig:hed_vs_hedprep}
\end{figure}

We first demonstrate the effect of our automatically constructed sketch domain (Section~\ref{sec:sketch-domain}). The key observation is that, while both raw HED edge maps~\cite{Xie_2015_ICCV} and our sketch domain lead to very similar conditional image completion, our sketch domain generalizes better to free hand sketches provided by users. Figure~\ref{fig:training_like_hed_vs_hedprep} compares conditional image completion based on automatically constructed (using the input image itself) HED edge maps and our sketch domain, showing that both lead to high quality results. In contrast, Figure~\ref{fig:hed_vs_hedprep} illustrates results from free hand user sketches. Synthesized results with HED (middle column) are blurrier and show some artifacts. Because the system is trained on highly accurate HED edge maps, artifacts in the user input translate into artifacts in the output. Our sketch domain leaves some ambiguity between image structures and sketch strokes, which enables the system to avoid artifacts when the user input is inaccurate.

\begin{figure}[t]
\centering
\includegraphics[width=1\linewidth]{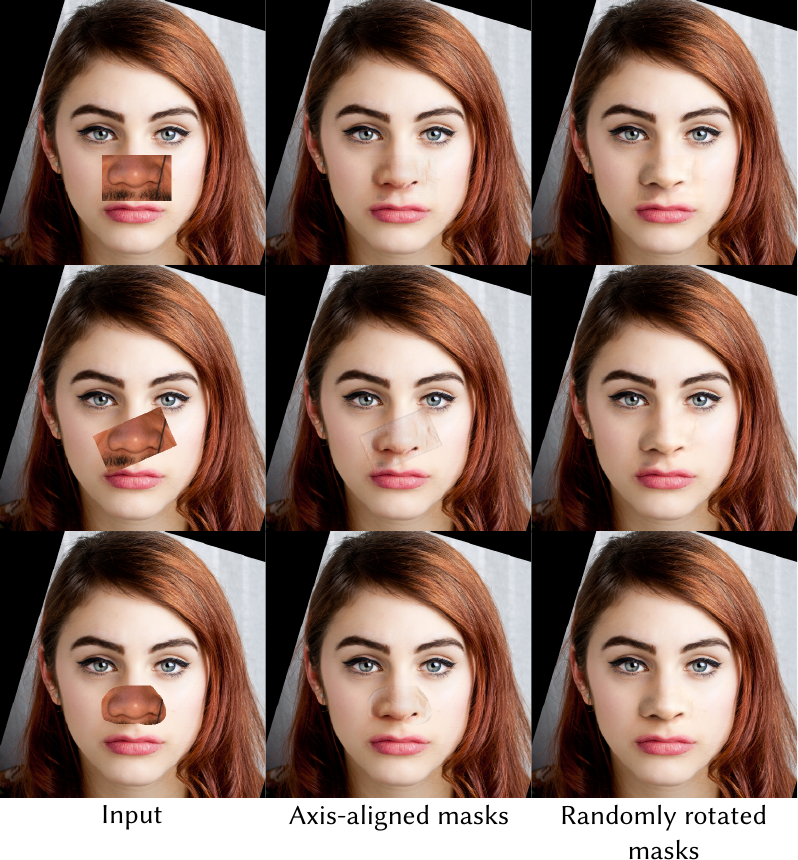}
\caption{Comparison of training with axis-aligned masks (middle) and randomly rotated masks (right). Randomly rotating masks avoids overfitting and generalizes to arbitrarily shaped masks. Photo courtesy of \href{https://www.flickr.com/photos/kaiteeteehee/8488585806/in/photolist-dW7dPG-qNRwp-dvq5VZ-qUwr3n-9zGefR-5xyG58-5DvBDC-nqx2Ww-nGQ9DS-qebpma-4XhXfT-4rBwhs-9KNvWX-8dZ7uM-buHDQ1-6397Zc-cWCHzh-7EAi7B-bxi821-5tpwpY-UvDkdX-rmWdZR-BncRrk-gAecuD-eLKEXD-D2Uzco-8WjiAx-6u8zLz-nGZMej-taYSw-aFobES-YpMTQc-4LVyfP-orRCox-eedrxA-7sUUHU-dRHaU2-8F6533-dyNC4N-Dez3q-XVF672-6BxaTH-aRbgQz-bk6KHN-94RkSc-9AYzzL-68AB4z-7TpKrD-hNeU7w-8McK9r}{Kaitee Silzer} and \href{https://www.flickr.com/photos/28781447@N04/28566305710/}{Patrick S}.}
\label{fig:rot_vs_norot}
\end{figure}

Figure~\ref{fig:rot_vs_norot} illustrates the benefit of using randomly rotated masks during training. Training only with axis aligned masks suffers from overfitting, and artifacts occur with non-axis aligned and arbitrarily shaped masks, as typically provided by users. On the other hand, our approach avoids these issues.

\begin{figure}[t]
\centering
\includegraphics[width=1\linewidth]{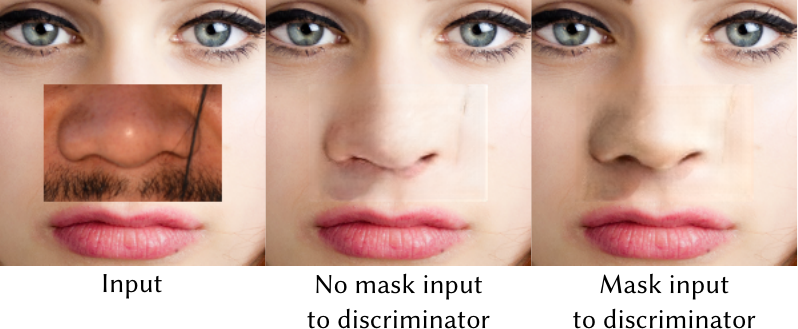}
\caption{Subtle artifacts appear if the mask is not provided to the discriminators (middle). Our approach feeds the mask to the discriminator and reduces these issues (right). Photo courtesy of \href{https://www.flickr.com/photos/kaiteeteehee/8488585806/in/photolist-dW7dPG-qNRwp-dvq5VZ-qUwr3n-9zGefR-5xyG58-5DvBDC-nqx2Ww-nGQ9DS-qebpma-4XhXfT-4rBwhs-9KNvWX-8dZ7uM-buHDQ1-6397Zc-cWCHzh-7EAi7B-bxi821-5tpwpY-UvDkdX-rmWdZR-BncRrk-gAecuD-eLKEXD-D2Uzco-8WjiAx-6u8zLz-nGZMej-taYSw-aFobES-YpMTQc-4LVyfP-orRCox-eedrxA-7sUUHU-dRHaU2-8F6533-dyNC4N-Dez3q-XVF672-6BxaTH-aRbgQz-bk6KHN-94RkSc-9AYzzL-68AB4z-7TpKrD-hNeU7w-8McK9r}{Kaitee Silzer} and \href{https://www.flickr.com/photos/28781447@N04/28566305710/}{Patrick S}.}
\label{fig:dismask_vs_nodismask}
\end{figure}

We demonstrate the need of including the mask in the input to the discriminators (see Figure~\ref{fig:dis_arch}) in Figure~\ref{fig:dismask_vs_nodismask}. Without providing the mask to the discriminator subtle artifacts around the mask boundaries occur, while our approach eliminates them.

\begin{figure}[t]
\centering
\includegraphics[width=1\linewidth]{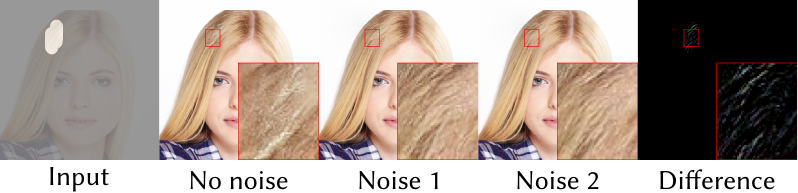}
\caption{Including a noise layer is important to produce realistic textures. Without it, synthesized textures exhibit artifacts (second image from left). We show results with two noise patterns (third and fourth image), and highlight the influence of the noise on texture details using a difference image (right). Photo from the public domain.}
\label{fig:noise}
\end{figure}

Finally, Figure~\ref{fig:noise} highlights the importance of using skip connections in the conditional generator (Figure~\ref{fig:gen_arch}) and a noise layer in its input. Without these provisions, the conditional completion network produces unrealistic textures (second image from the left). In contrast, with our approach we obtain a high quality output (third and fourth image). We show two results with different noise patterns to emphasize the influence of the noise on the texture details, visualized by the difference image on the right. 

\subsection{Sketch-based Face Image Editing}
\label{sec:sketch-based-face-image-editing}

\begin{figure*}[t]
\centering
\includegraphics[width=1\textwidth]{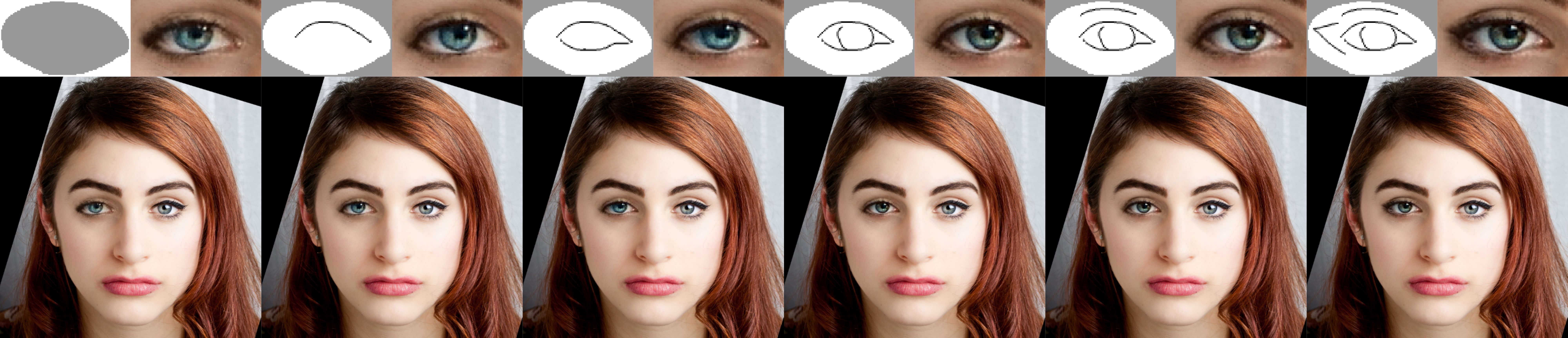}
\caption{Progressive addition of strokes. Void sketch input (first column) produces artifacts, although the output is semantically correct. Adding only a few strokes resolves most of the ambiguities and results in plausible renderings (columns two and three). By adding more strokes, our system enables to indicate fine details that translate into realistic results in a predictable manner (remaining columns). Photo courtesy of \href{https://www.flickr.com/photos/kaiteeteehee/8488585806/in/photolist-dW7dPG-qNRwp-dvq5VZ-qUwr3n-9zGefR-5xyG58-5DvBDC-nqx2Ww-nGQ9DS-qebpma-4XhXfT-4rBwhs-9KNvWX-8dZ7uM-buHDQ1-6397Zc-cWCHzh-7EAi7B-bxi821-5tpwpY-UvDkdX-rmWdZR-BncRrk-gAecuD-eLKEXD-D2Uzco-8WjiAx-6u8zLz-nGZMej-taYSw-aFobES-YpMTQc-4LVyfP-orRCox-eedrxA-7sUUHU-dRHaU2-8F6533-dyNC4N-Dez3q-XVF672-6BxaTH-aRbgQz-bk6KHN-94RkSc-9AYzzL-68AB4z-7TpKrD-hNeU7w-8McK9r}{Kaitee Silzer}.}
\label{fig:progressive}
\end{figure*}

Figure~\ref{fig:progressive} shows outputs by our system when proressively adding input strokes. Once the user provides a few strokes, the system renders plausible results that can be refined intuitively by indicating fine details with additional strokes. We refer to the supplemental video for more interactive results.

\begin{figure*}[t]
\centering
\includegraphics[width=1\textwidth]{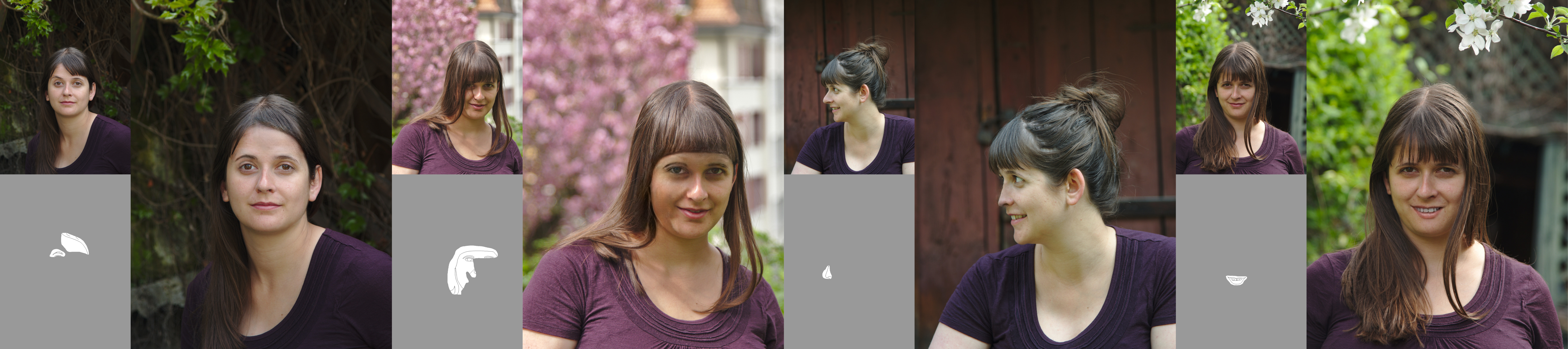}
\caption{Editing results on ``in the wild'' images captured using a DSLR camera. Our system renders high-quality results even for particularly difficult examples, such as replacing a substantial part of the face or profile portraits.}
\label{fig:in_the_wild}
\end{figure*}

In Figure~\ref{fig:in_the_wild} we apply our system to ``in the wild'' images. For this purpose, we shot some portrait images using a DSLR camera. The system renders high-quality results even for particularly difficult examples, such as replacing a substantial part of the face or profile portraits.

\begin{figure*}[t]
\centering
\includegraphics[width=0.96\textwidth]{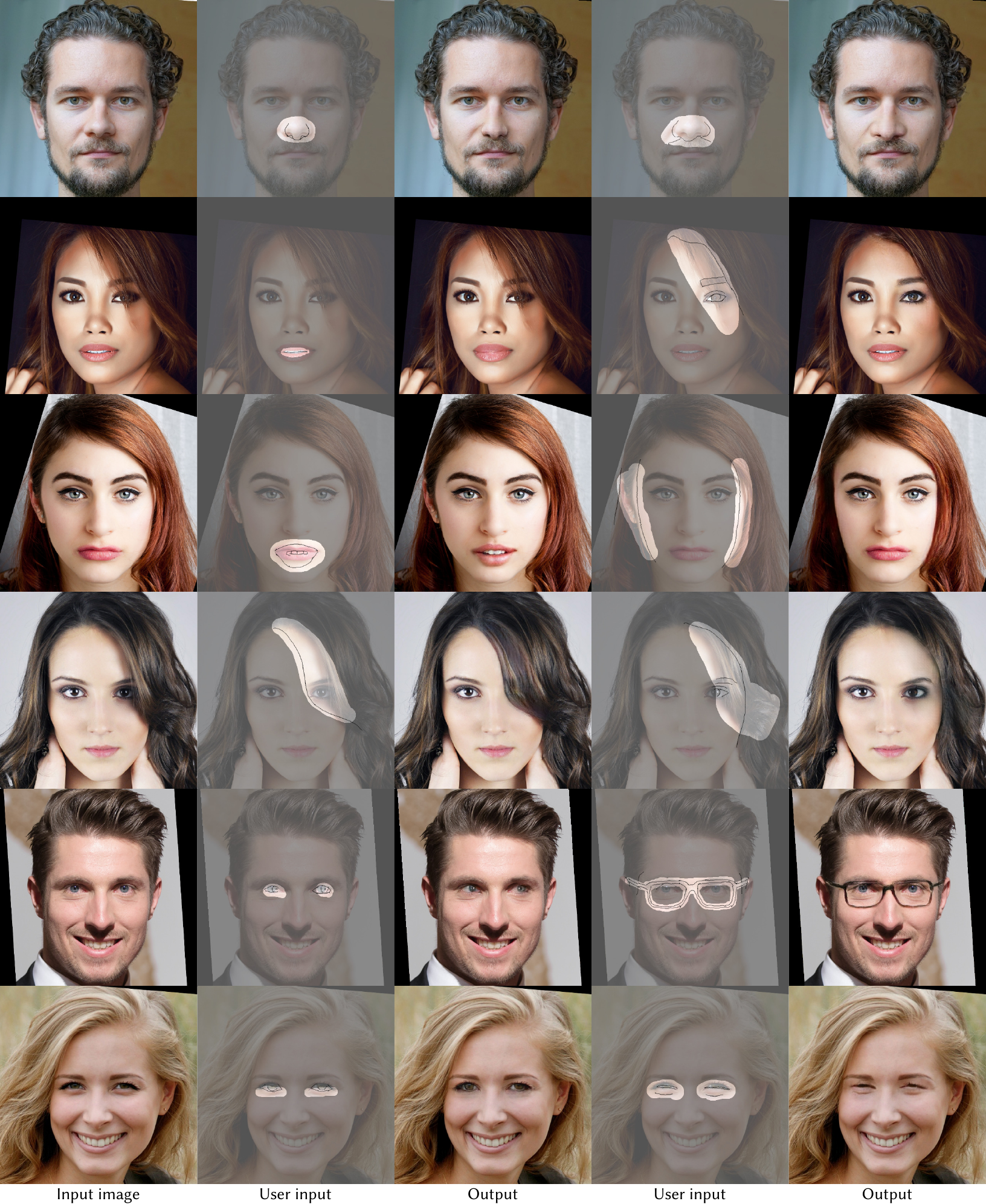}
\caption{Results using sketching. The examples shown here include editing of nose shape, facial expression (open or close mouth), hairstyle, eyes (open or close eyes, gaze direction, glasses), and face shape. Synthesized face regions exhibit plausible shading and texture detail. Photo courtesy of \href{https://www.flickr.com/photos/kaiteeteehee/8488585806/in/photolist-dW7dPG-qNRwp-dvq5VZ-qUwr3n-9zGefR-5xyG58-5DvBDC-nqx2Ww-nGQ9DS-qebpma-4XhXfT-4rBwhs-9KNvWX-8dZ7uM-buHDQ1-6397Zc-cWCHzh-7EAi7B-bxi821-5tpwpY-UvDkdX-rmWdZR-BncRrk-gAecuD-eLKEXD-D2Uzco-8WjiAx-6u8zLz-nGZMej-taYSw-aFobES-YpMTQc-4LVyfP-orRCox-eedrxA-7sUUHU-dRHaU2-8F6533-dyNC4N-Dez3q-XVF672-6BxaTH-aRbgQz-bk6KHN-94RkSc-9AYzzL-68AB4z-7TpKrD-hNeU7w-8McK9r}{Kaitee Silzer} and \href{https://commons.wikimedia.org/wiki/File:Marcel_Hirscher_Laura_Moisl_Gala_Nacht_des_Sports_Österreich_2015.jpg}{Manfred Werner} (third, fifth row), other photos from the public domain.}
\label{fig:sketch_results}
\end{figure*}

Figure~\ref{fig:sketch_results} shows various results obtained with our system using sketching. They include editing operations to change the nose shape, facial expression (open or close mouth), hairstyle, eyes (open or close eyes, gaze direction, glasses), and face shape. These examples demonstrate that by providing simple sketches, users can obtain predictive, high quality results with our system. Edges in the user sketches intuitively determine the output, while the system is robust to small imperfections in the sketches, largely avoiding any visible artifacts in the output. Synthesized face regions exhibit plausible shading and texture detail. In addition, these results underline the flexibility of our system, which is able to realize a variety of intended user edits without being restricted to specific operations (like opening eyes~\cite{dolhansky2017eye}).

\begin{figure*}[t]
\centering
\includegraphics[width=0.96\textwidth]{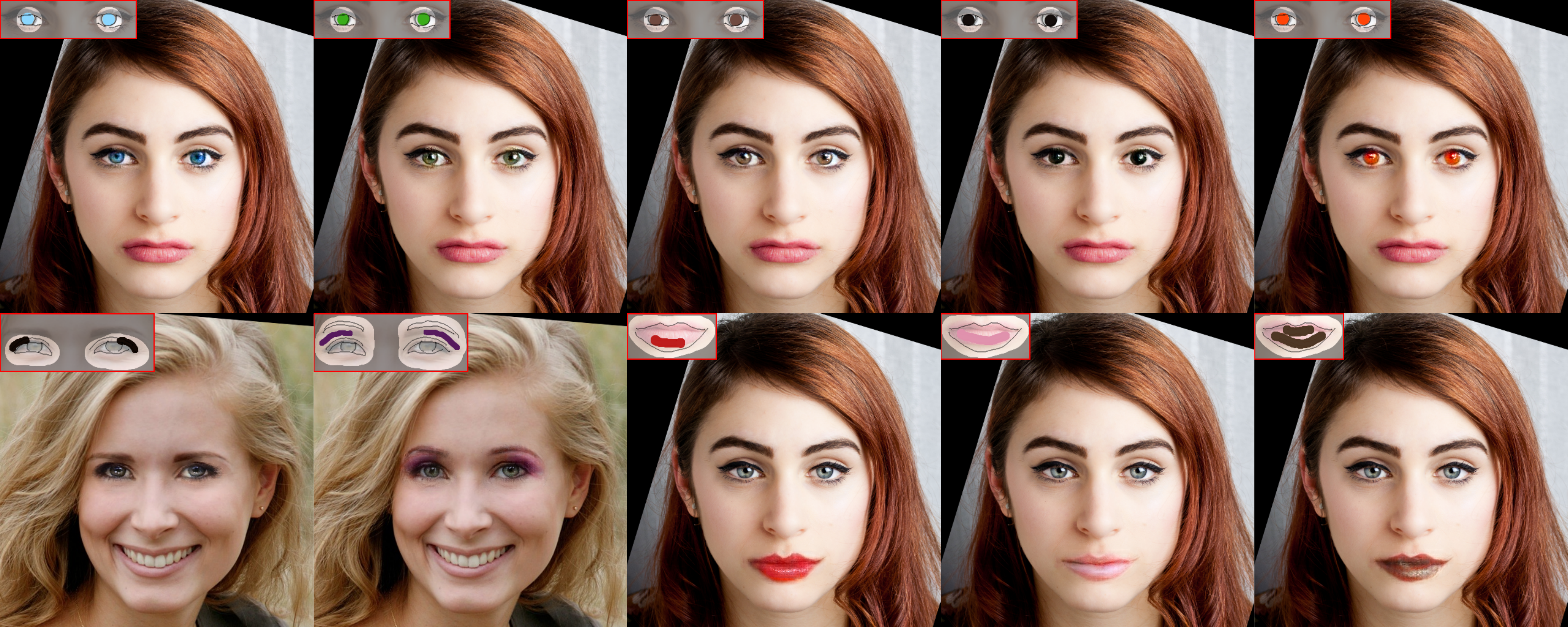}
\caption{Results using sketching and coloring. The unmodified input images are shown in Figure~\ref{fig:sketch_results} in rows three and six, leftmost column. Photo courtesy of \href{https://www.flickr.com/photos/kaiteeteehee/8488585806/in/photolist-dW7dPG-qNRwp-dvq5VZ-qUwr3n-9zGefR-5xyG58-5DvBDC-nqx2Ww-nGQ9DS-qebpma-4XhXfT-4rBwhs-9KNvWX-8dZ7uM-buHDQ1-6397Zc-cWCHzh-7EAi7B-bxi821-5tpwpY-UvDkdX-rmWdZR-BncRrk-gAecuD-eLKEXD-D2Uzco-8WjiAx-6u8zLz-nGZMej-taYSw-aFobES-YpMTQc-4LVyfP-orRCox-eedrxA-7sUUHU-dRHaU2-8F6533-dyNC4N-Dez3q-XVF672-6BxaTH-aRbgQz-bk6KHN-94RkSc-9AYzzL-68AB4z-7TpKrD-hNeU7w-8McK9r}{Kaitee Silzer}, other photo from the public domain.}
\label{fig:coloring_results}
\end{figure*}

Figure~\ref{fig:coloring_results} highlights the color editing functionality of our system, where we show examples of changing eye, makeup, and lip color. This shows that the system is robust to imperfect user input consisting of only rough color strokes, largely avoiding artifacts. 

\subsection{Smart Copy-Paste}
\label{sec:smart-copy-paste}

\begin{figure*}[t]
\centering
\includegraphics[width=0.98\textwidth]{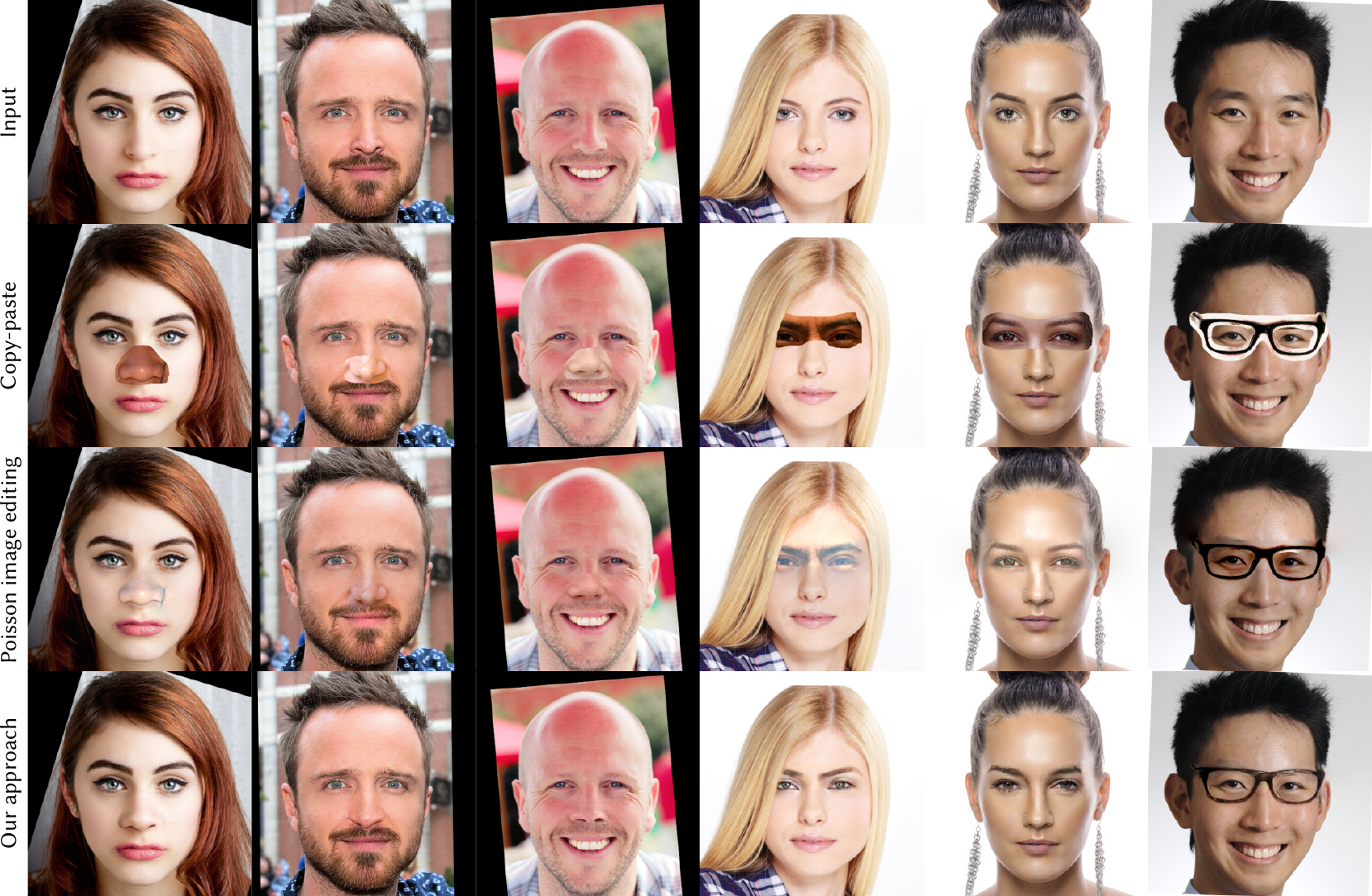}
\caption{Smart copy-paste results. Different rows show input images, copy-paste input, results from Poisson image editing, and our approach. Poisson image editing copies gradients, while we copy the sketch domain of the source region. While our results are mostly free of artifacts, Poisson image editing suffers from inconsistencies between source and target. Photo courtesy of \href{https://www.flickr.com/photos/kaiteeteehee/8488585806/in/photolist-dW7dPG-qNRwp-dvq5VZ-qUwr3n-9zGefR-5xyG58-5DvBDC-nqx2Ww-nGQ9DS-qebpma-4XhXfT-4rBwhs-9KNvWX-8dZ7uM-buHDQ1-6397Zc-cWCHzh-7EAi7B-bxi821-5tpwpY-UvDkdX-rmWdZR-BncRrk-gAecuD-eLKEXD-D2Uzco-8WjiAx-6u8zLz-nGZMej-taYSw-aFobES-YpMTQc-4LVyfP-orRCox-eedrxA-7sUUHU-dRHaU2-8F6533-dyNC4N-Dez3q-XVF672-6BxaTH-aRbgQz-bk6KHN-94RkSc-9AYzzL-68AB4z-7TpKrD-hNeU7w-8McK9r}{Kaitee Silzer}, \href{https://www.flickr.com/photos/28781447@N04/28566305710/}{Patrick S}, \href{https://www.flickr.com/photos/gdcgraphics/8023002250/}{Gordon Correll}, \href{http://www.norden.org/en/news-and-events/images/people/others/pia-leppaelae/view}{NN norden.org}, \href{https://commons.wikimedia.org/wiki/File:George_Clarke_(filmmaker).jpg}{Kennyjmuk}, and \href{https://www.flickr.com/photos/time-to-look/24207775273/in/photolist-23on75w-HBfCWY-qocC4Z-23u47dH-CTabcK-qrRfdM-qRC7nE-qcGiLe-75BqHp-63E1dM-5uFzXJ-5LLSpx-x5A2Zd-vwZv5Z-pVUVZf-gbykP8-F9bEhc-oKA7CZ-ozKKS2-9nMMgc-51bBHc-5LwPVq-Cufguc-p5YtVu-v8x51K-wpWxGm}{Ted McGrath}, other photos from the public domain.}
\label{fig:cp_vs_poisson}
\end{figure*}

Figure~\ref{fig:cp_vs_poisson} shows results of our smart copy-paste functionality, where we copy the masked sketch domain of a source image into a target image. This allows users to edit images in an example-based manner, without the need to draw free hand sketches. In addition, we compare our results to Poisson image editing~\cite{Perez2003}. The comparison reveals that Poisson image editing often produces artifacts due to mismatches in shading (first, fourth, fifth, sixth column) or face geometry (second, third, fifth column) between the source and target images. Our approach is more robust and avoids artifacts, although significant misalignment between source and target still leads to unnatural outputs (third column).

\subsection{Comparisons to Related Techniques}
\label{sec:comparisons}
In this section we compare our system to unconditional face completion and image translation. Additional results can be found in the supplemental material.

\subsubsection{Unconditional Image Completion}
\label{sec:comparison_iizuka}

\begin{figure*}[t]
\centering
\includegraphics[width=1\textwidth]{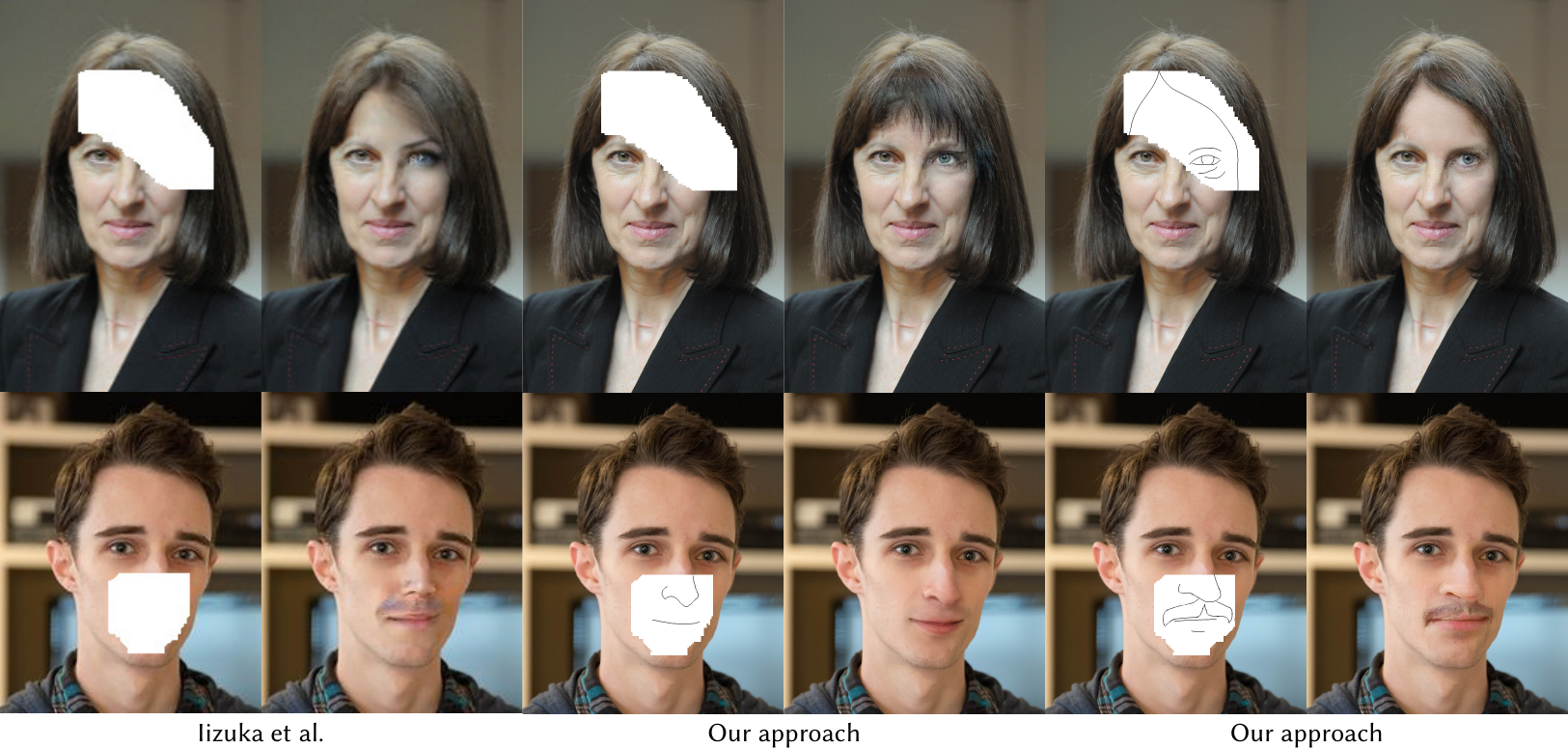}
\caption{Comparison to~\cite{IizukaSIGGRAPH2017}. In contrast to Iizuka et al., our system is able to render multiple different results for completing the same image region. Note that our results are of much higher resolution and suffer less from visual artifacts. Additional results can be found in the supplemental material. Photo courtesy of Satoshi Iizuka~\shortcite{IizukaSIGGRAPH2017} and \href{https://www.flickr.com/photos/ctcwired/24806065080/}{Paradox Wolf}.}
\label{fig:comparison_iizuka}
\end{figure*}

The image completion system by Iizuka et al.~\shortcite{IizukaSIGGRAPH2017} was trained on face images, which enables us to perform a comparison to our approach. For this purpose, we copy results published in their paper and apply our system on these examples. For each example, we use the corresponding high-resolution version of the image for our approach. Figure~\ref{fig:comparison_iizuka} shows the results produced using our system (fourth and sixth column) in comparison to the results published by Iizuka et al. (second column). The main conceptual difference is that our system can render various results for completing the same image region by providing different sketch input. Moreover, the outputs of our network are mostly free of artifacts and of much higher resolution than the results published by Iizuka et al.

\subsubsection{Image Translation}
\label{sec:comparison_isola}

\begin{figure*}[t]
\centering
\includegraphics[width=1\textwidth]{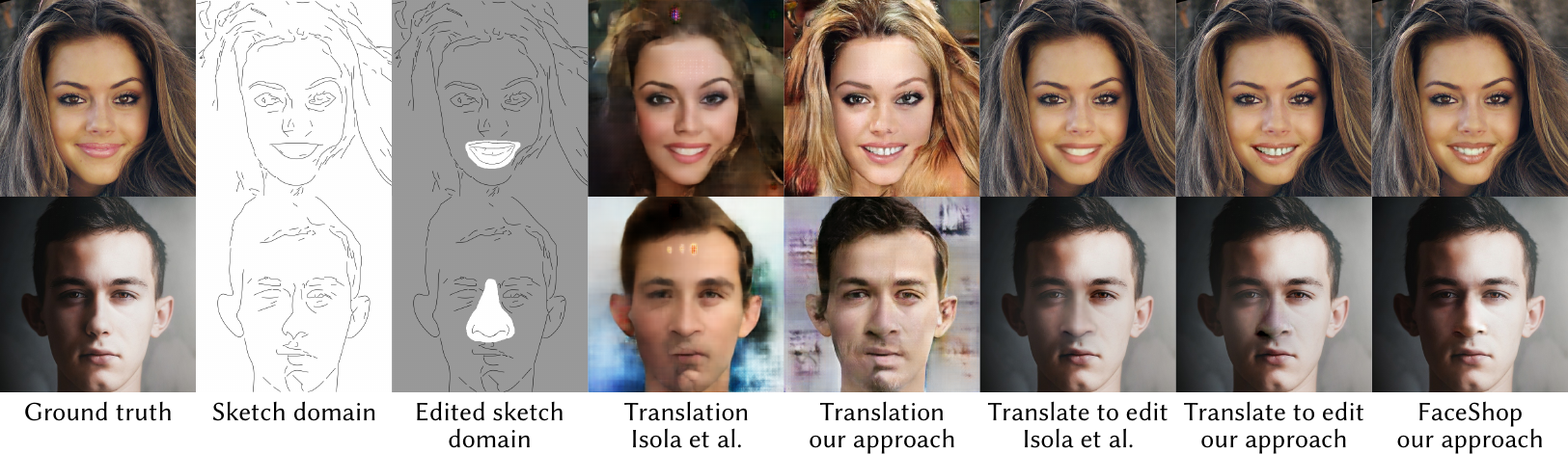}
\caption{Comparison to~\cite{Isola_2017_CVPR}. Our network, trained on image translation (column five), produces more detailed textures and less artifacts compared to Isola et al. (fourth column). Moreover, our proposed approach (last column) leads to more consistent results than the ``translate to edit'' approaches (see Section~\ref{sec:comparison_isola}). Additional results can be found in the supplemental material. Photos from the public domain.}
\label{fig:comparison_isola}
\end{figure*}

An image translation network, trained to translate from sketches to photos can be used to perform local image edits in a ``translate to edit'' manner as follows: an input image is first mapped to the sketch domain. Next, the resulting sketch can be edited locally and further translated to render a photo. Finally, the edited region can be blended with the input image using Poisson image editing~\cite{Perez2003}. To evaluate the performance of this approach, we train the pix2pix network~\cite{Isola_2017_CVPR} to translate our sketch domain to photos. To emulate our editing approach, we train with both sketch and color domain inputs. For comparison, we also train our proposed network on the task of image translation. In this case, the local discriminator gets a random crop as input. In Figure~\ref{fig:comparison_isola} we compare our network, trained on image translation, to pix2pix, including the ``translate to edit'' approach, and our proposed conditional completion technique. The comparison shows that our network produces translations with less artifacts and more detailed textures, compared to pix2pix. Moreover, the comparison shows that the ``translate to edit'' approach can work for simple edits (first row), but generally produces worse results than our proposed conditional completion framework. First, the problem of translating an entire image is much more ambiguous than translating only an image region, resulting in lower quality renderings. Second, much information is lost due to the mapping to the sketch domain, resulting in inconsistent results, such as inappropriate shading due to missing shadows. In contrast, our proposed system is robust to these issues and produces high-quality and consistent results.

\subsection{Limitations}

\begin{figure*}[t]
\centering
\includegraphics[width=1\textwidth]{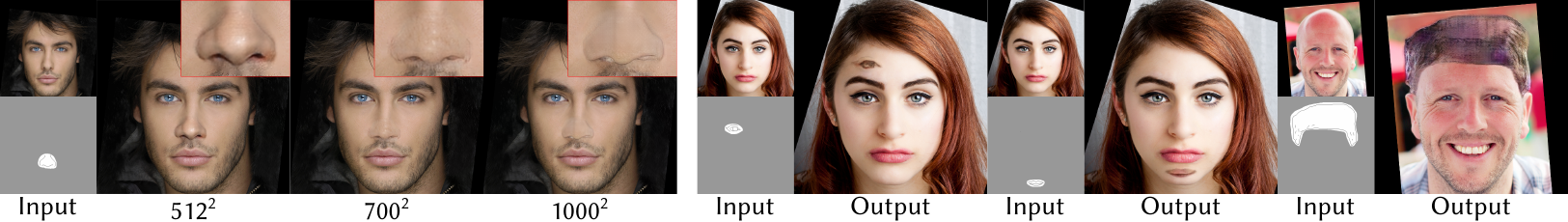}
\caption{Left: generalization capability to higher resolutions. Our system generalizes to some extent to higher resolutions, but fails at significantly higher resolutions. Typical failure cases of our system are shown on the right. Semantically inconsistent content like an eye on the forehead or a mouth on the chin confuse the completion network. Another failure is to completely change hairstyle. Photo courtesy of \href{https://commons.wikimedia.org/wiki/File:MARTAKIS1.jpg}{Bratopoulosm}, \href{https://www.flickr.com/photos/kaiteeteehee/8488585806/in/photolist-dW7dPG-qNRwp-dvq5VZ-qUwr3n-9zGefR-5xyG58-5DvBDC-nqx2Ww-nGQ9DS-qebpma-4XhXfT-4rBwhs-9KNvWX-8dZ7uM-buHDQ1-6397Zc-cWCHzh-7EAi7B-bxi821-5tpwpY-UvDkdX-rmWdZR-BncRrk-gAecuD-eLKEXD-D2Uzco-8WjiAx-6u8zLz-nGZMej-taYSw-aFobES-YpMTQc-4LVyfP-orRCox-eedrxA-7sUUHU-dRHaU2-8F6533-dyNC4N-Dez3q-XVF672-6BxaTH-aRbgQz-bk6KHN-94RkSc-9AYzzL-68AB4z-7TpKrD-hNeU7w-8McK9r}{Kaitee Silzer}, and \href{https://commons.wikimedia.org/wiki/File:George_Clarke_(filmmaker).jpg}{Kennyjmuk}.}
\label{fig:limitations}
\end{figure*}

Figure~\ref{fig:limitations} (left) shows the generalization capability of our system to higher resolutions. The system generalizes to some extent to higher resolutions, but fails to render structures at significantly higher resolutions, since it was not trained on structures at these scales. Figure~\ref{fig:limitations} (right) shows typical failure cases of our system. During training, the network never saw semantically inconsistent layouts like an eye on the forehead or a mouth on the chin. Another failure case is to completely change hairstyle. Our system never had to invent hair texture during training, it only learned to copy hair texture from the context to moderately modify the hairstyle.

\section{Conclusions}

In this paper we presented a novel system for sketch-based image editing based on generative adversarial neural networks (GANs). We introduce a conditional completion network that synthesizes realistic image regions based on user input consisting of masks, simple edge sketches, and rough color strokes. In addition, our approach also supports example-based editing by copying and pasting source regions into target images. We train our system on high resolution imagery based on the celebA face dataset and show a variety of successful and realistic editing results.  Key to our method is a careful design of the training data, the architectures of our conditional generator and discriminator networks, and the loss functions. In particular, we describe suitable automatic construction of sketch and color stroke training data. In the future, we will train our system on more diverse and even higher resolution image datasets. We are also exploring additional user interaction tools that may enable even more intuitive editing workflows. Finally, we will explore how to leverage deep learning for sketch-based editing of video data.

\bibliographystyle{ACM-Reference-Format}
\bibliography{bibliography}

\end{document}